# WAVELET BASED IMAGE CODING SCHEMES: A RECENT SURVEY

Rehna V. J[1] and Jeya Kumar M. K[2]

[1]Department of Electronics & Communication Engineering, Noorul Islam University, Kumarakoil, Tamil Nadu
`rehna_vj@yahoo.co.in`
[2]Department of Computer Applications, NIU, Kumarakoil, Tamil Nadu
`jeyakumarmk@yahoo.com`

*ABSTRACT*

*A variety of new and powerful algorithms have been developed for image compression over the years. Among them the wavelet-based image compression schemes have gained much popularity due to their overlapping nature which reduces the blocking artifacts that are common phenomena in JPEG compression and multiresolution character which leads to superior energy compaction with high quality reconstructed images. This paper provides a detailed survey on some of the popular wavelet coding techniques such as the Embedded Zerotree Wavelet (EZW) coding, Set Partitioning in Hierarchical Tree (SPIHT) coding, the Set Partitioned Embedded Block (SPECK) Coder, and the Embedded Block Coding with Optimized Truncation (EBCOT) algorithm. Other wavelet-based coding techniques like the Wavelet Difference Reduction (WDR) and the Adaptive Scanned Wavelet Difference Reduction (ASWDR) algorithms, the Space Frequency Quantization (SFQ) algorithm, the Embedded Predictive Wavelet Image Coder (EPWIC), Compression with Reversible Embedded Wavelet (CREW), the Stack-Run (SR) coding and the recent Geometric Wavelet (GW) coding are also discussed. Based on the review, recommendations and discussions are presented for algorithm development and implementation.*

## 1. INTRODUCTION

Today's digital world deals with huge amounts of digital data (audio, image & video). Despite the rapid progress in system performance, processor speed and mass storage density; demand for greater digital data storage capacity and transmission bandwidth is overwhelming. Data compression [1], particularly image compression [2] plays a very vital role in the field of multimedia computer services and other telecommunication applications. The field of image compression has a wide spectrum of methods ranging from classical lossless techniques and popular transform approaches to the more recent segmentation based (or second generation) coding methods. Until recently, the discrete cosine transform (DCT) [8] has been the most popular technique for compression due to its optimal performance and ability to be implemented at a reasonable cost. Several commercially successful compression algorithms, such as the JPEG standard [3] for still images and the MPEG standard [4] for video images are based on DCT. Wavelet-based techniques [5] are the latest development in the field of image compression which offers multiresolution capability leading to superior energy compaction with high quality

                                                                                                               101

reconstructed images at high compression ratios. The wavelet transform has emerged as a cutting edge technology, within the field of image compression. The DWT has taken over DCT [7] due to its ability to solve the problem of blocking artifacts introduced during DCT compression. It also reduces the correlation between the neighboring pixels and gives multi scale sparse representation of the image. Wavelet compression provides excellent results in terms of rate-distortion compression. Over the past few years, a variety of powerful and sophisticated wavelet-based schemes for image compression have been developed and implemented. The new ISO/ITU-T standard for still image coding, JPEG 2000, is a wavelet-based compression algorithm. This second generation algorithm is being designed to address the requirements of different kinds of applications including internet, digital photography, image databases, document imaging, remote sensing, scanning, laser printing, facsimile, mobile applications, medical imagery, digital library etc.

The paper is organized in the following way. Section II deals with the basics of image compression. A discussion of performance measures of different algorithms is presented in section III. A brief introduction on wavelets for image compression is described in section IV. Section V gives an outline of various state of the art wavelet-based coding schemes. The salient features of each of these schemes are discussed in later sub-sections. The comparison of different wavelet based methods is given in section VI as a summary and the paper is concluded in section VII.

## 2. IMAGE COMPRESSION BASICS

The neighbouring pixels of most natural images are highly correlated and thus contain lot of redundant information. A less correlated representation of the image is the result of any compression algorithm. The main task of image compression algorithms is reduction of redundant and irrelevant information. In the present scenario, various methods of compressing still images exist. In any data compression scheme, three basic steps are involved: Transformation, Quantization and Encoding. The basic block diagram of an image compression system is shown in Figure 1.

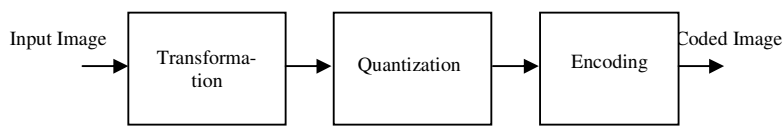

Figure 1. Elements of an Image Compression System

### 2.1. Transformation

In image compression, transform is indented to de-correlate the input pixels. Selection of proper transform is one of the important issues in image compression schemes. The transform should be selected in such a way that it reduces the size of the resultant data set as compared to the source data set. Few transformations reduce the numerical size of the data items that allow them to represent by fewer binary bits. The technical name given to these methods of transformation is mapping [28]. Some mathematical transformations have been invented for the sole purpose of data compression; others have been borrowed from various applications and applied to data compression. These include the Discrete Fourier Transform (DFT), Discrete Cosine Transform (DCT) [30], Walsh-Hadamard Transform (WHT), Hadamard-Haar Transform (HHT), Karhune-Loeve Transforms (KLT), Slant-Haar Transform (SHT), Short Fourier Transforms (SFT), and Wavelet Transforms (WT) [10]. Transform selection process still remains an active field of research.

## 2.2. Quantization

The procedure of approximating the continuous set of values in the image data with a finite, preferably small set of values is called quantization. The original data is the input to a quantizer and the output is always one among a limited number of levels. The quantization step may also be followed by the process of thresholding. Each sample is scaled by a quantization factor in the process of quantization, whereas the samples are eliminated if the value of the sample is less than the defined threshold value, in the process of thresholding. These two methods are responsible for the introduction of error and leads to degradation of quality. The degradation is based on selection of quantization factor and the value of threshold. If the threshold value is high, the loss of information is more, and vice versa. The value of threshold or the quantization factor should be selected in a way that it satisfies the constraints of human visual system for better visual quality at high compression ratios. Quantization is a process of approximation. A good quantizer is the one which represents the original signal with minimum distortion. If lossless compression is desired this step should be eliminated.

## 2.3. Encoding

Encoding process reduces the overall number of bits required to represent the image [1]. An entropy encoder compresses the quantized values further to give better overall compression. This process removes the redundancy in the form of repetitive bit patterns at the output of the quantizer. It uses a model to precisely determine the probabilities for each quantized value and produces a suitable code based on these probabilities so that the resultant output code stream will be smaller than the input. Commonly used entropy coders are the Huffman encoder and the Arithmetic encoder. The Huffman procedure needs each code to have an integral number of bits, while arithmetic coding techniques allow for fractional number of bits per code by grouping two or more similar codes together into a block composed of an integral number of bits. This makes arithmetic codes to perform better than Huffman codes. Therefore, arithmetic codes are more commonly used in wavelet based algorithms. The decoding process involves the reverse operation of the encoding steps with the exception of de-quantization step that cannot be reversed exactly.

## 3. PERFORMANCE MEASURES

There are a variety of ways in which different compression algorithms can be evaluated and compared. For quantifying the error between images, two measures are being commonly used. They are Mean Square Error (MSE) and Peak Signal to Noise Ratio (PSNR). The MSE between two images f and g, where f represents the original image and g represents the reconstructed image, is defined as:

$$MSE = 1/N \sum_{x,y} [f(x,y) - g(x,y)]^2 \qquad (1)$$

where N is the total number of pixels in each image and the sum over x,y denotes the sum over all pixels in the image. The PSNR between two (8 bpp) images, in decibels is given by:

$$PSNR = 10 \log_{10} \left( \frac{255^2}{MSE} \right) \qquad (2)$$

PSNR is used more often, since it is a logarithmic measure, and human brains seem to respond logarithmically to intensity. Increasing PSNR means increasing fidelity of compression. Generally, when the PSNR is greater than or equal to 40 dB, it is said that the two images are virtually indistinguishable by human observers.

## 4. WAVELETS FOR IMAGE COMPRESSION

Among a variety of new and powerful algorithms that have been developed for image compression over the years, the wavelet-based image compression has gained much popularity due to their overlapping nature which reduces the blocking artifacts and multiresolution character, leading to superior energy compaction with high quality reconstructed images. Wavelet-based coding [6] provides substantial improvements in picture quality at higher compression ratios. Furthermore, at higher compression ratios, wavelet coding methods degrade much more gracefully than the block-DCT methods. Since the wavelet basis consists of functions both with short support and long support for high frequencies and for low frequencies respectively, smooth areas of the image may be represented with very few bits, and detail can be added where ever required [8]. Their superior energy compaction properties and correspondence with the human visual system have made, wavelet compression methods produce subjective results [1]. Due to the many advantages, wavelet based compression algorithms have paved way for the for the new JPEG-2000 standard [3].

## 5. WAVELET BASED CODING SCHEMES

Wavelet compression schemes allow the integration of various compression techniques into one. With wavelets, a compression ratio of up to 300:1 is achievable [29]. A number of novel and sophisticated wavelet-based schemes for image compression have been developed and implemented over the past few years. Some of the most popular schemes are discussed in the paper. These include Embedded Zero Tree Wavelet (EZW) [9], Set-Partitioning in Hierarchical Trees (SPIHT) [11], Set Partitioned Embedded Block Coder (SPECK) [12], Embedded Block Coding with Optimized Truncation (EBCOT) [13], Wavelet Difference Reduction (WDR )[14], Adaptively Scanned Wavelet Difference Reduction (ASWDR) [15] , Space – Frequency Quantization (SFQ) [16], Embedded Predictive Wavelet Image Coder (EPWIC) [18], Compression with Reversible Embedded Wavelet (CREW) [19], the Stack- Run (SR) [20], the recent Geometric Wavelet (GW) [21 ] and improved GW [22 ]. Each of these 'state of the art' algorithms are dealt and discussed in detail in the sections below.

### 5.1. The EZW (Embedded Zerotree Wavelet) Coding

Image coding techniques exploiting scalar quantization on hierarchical structures of transformed images have been very effective and computationally less complicated. The EZW coding for image compression is a simple, remarkably effective algorithm of this kind introduced by Shapiro [9] in his paper "Embedded Image Coding using Zerotrees of Wavelet Coefficients" in 1993. This work was one of the first to show the supremacy of wavelet based techniques for image compression. EZW coding exploits the multi resolution properties of the wavelet transforms [27] to give a computationally simple algorithm with better performance compared to other existing wavelet transforms.

An embedded coding is a process of encoding the transform magnitudes which allow progressive transmission of the compressed image. Embedded encoding is also called progressive encoding. The Embedded Zerotree Wavelet encoder is based on progressive coding to compress an image into a bit stream with increasing accuracy. This means that as more bits are added to the stream, the decoded image will contain more detail. The encoding method used by the EZW is the bit-plane encoding.

Zerotrees are a concept that allows for a concise encoding of the positions of significant values that result during the embedded coding process. The EZW encoding is based on two main observations:

1. When the wavelet transformation of an image is taken, the energy in the sub bands decreases as the scale decreases (low scale means high resolution). So on an average, the wavelet coefficients will be smaller in the higher sub bands than in the lower sub bands. The higher sub bands add only detail. Since natural images generally have a low pass spectrum, progressive encoding is a very natural choice for compressing wavelet transformed images.

2. Large wavelet coefficients are more vital than the small ones.

These interpretations are exploited by encoding the wavelet coefficients in decreasing order and in several passes. For every pass, a threshold value is chosen against which all the wavelet coefficients are measured. A wavelet coefficient is encoded and removed from the image if it is larger than the threshold. If the wavelet coefficient is smaller, it is left for the next pass. When all the coefficients have been scanned, the threshold is lowered and the image is scanned again to add more detail to the already encoded image. This process is repeated until some added criterion has been satisfied or all the wavelet coefficients have been completely encoded.

A detailed review of the EZW coding is presented here. A concrete understanding of this algorithm is necessary to make it much easier to comprehend the other algorithms discussed in later sections which are built upon the fundamental concepts of EZW.

A zerotree is a quadtree which has insignificant wavelet transform values at each of its locations, for a given threshold T0. This quad-tree has all nodes that are equal to or smaller than the root. The tree is encoded with a single symbol and it is then reconstructed by the decoder as a quad-tree filled with zeros. The root has to be smaller than the threshold value against which the wavelet coefficients are measured. The EZW encoder exploits the zerotree based on the observation that the wavelet coefficients decrease with scale. The zerotree is based on the supposition that if a wavelet coefficient at a coarse scale is insignificant with respect to a given threshold T0, then all wavelet coefficients of the same orientation in the same spatial location at a finer scales are likely to be insignificant with respect to the threshold T0. The idea is to define a tree of zero symbols which starts at a root that is also zero and labeled as end-of-block. Zerotrees can only be useful if they occur frequently. Fortunately, the multiresolution structures of the wavelet transform of natural scenes do produce many zerotrees (especially at higher thresholds).
The EZW algorithm encodes the so obtained tree structure. This results in bits that are generated in the order of importance, leading to a fully embedded code. The core advantage of this encoding is that the encoder can terminate the encoding at any point, by this means allowing a target bit rate to be met exactly. The process is repeated after lowering the threshold, till the threshold has become smaller than the smallest coefficient to be transmitted. This is done to arrive at a perfect reconstruction. Likewise, the decoder can also stop decoding at any point of time, resulting in an image that would have been produced at the rate of the truncated bit stream. A major concern is on the transmission of the coefficient positions. Without this information the decoder will not be able to reconstruct the original signal (although it can perfectly reconstruct the transmitted bit stream). It is in the encoding of the coefficient positions where the efficient encoders are separated from the inefficient ones.

A predefined scan order is used to encode the position of the wavelet coefficients in EZW coding. Many positions are encoded implicitly through the use of zero-trees. Several scan orders are accessible, provided that the lower sub bands are completely scanned before going on to the higher sub bands. The scan order, as shown in Figure 2, is the most commonly used.

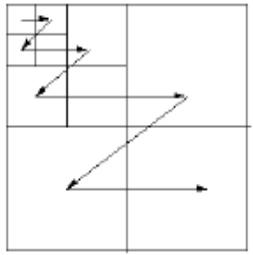

Figure 2. Scan order used in EZW coding

The algorithm gives brilliant results without any training, pre-stored tables or codebooks, or prior knowledge of the image source. EZW encoder does not actually compress anything; it only reorders the wavelet coefficients in such a way that they can be compressed very efficiently. Table 1 shows the results of compression ratios and PSNR values for Lena test image of size 512 X 512 using EZW algorithm.

Table 1. Results for Lena test image of size 512x512:
Compression Ratio & PSNR using EZW

| Bit Rate | Compression Ratio | MSE | PSNR |
|---|---|---|---|
| 1.0 | 8 | 7.21 | 39.55 |
| 0.5 | 16 | 15.32 | 36.28 |
| 0.25 | 32 | 31.33 | 33.17 |
| 0.125 | 64 | 61.67 | 30.23 |
| 0.0625 | 128 | 114.5 | 27.54 |
| 0.03125 | 256 | 188.27 | 25.38 |

The artifacts produced at low bit rates using this method are unavoidable and are the common characteristics of all wavelet coding schemes coded to the same PSNRs. However, these artifacts are subjectively not objectionable like the blocking effects produced by DCT [29] or other block transform based coding schemes. The EZW encoder is always followed by a symbol encoder, for example an arithmetic encoder, due to the above mentioned reason.

The performance measure of EZW is used as reference for comparison with the new techniques of image compression. Hence, it has become one of the state-of-the-art algorithms for image compression.

### 5.2. The SPIHT (Set Partitioning in Hierarchical Trees) Coding

The SPIHT coding [11] is an improved version of the EZW algorithm that achieves higher compression and better performance than EZW. It was introduced by Said and Pearlman [25] in 1996. SPIHT is expanded as Set Partitioning in Hierarchical Trees. The term Hierarchical Trees refers to the quadtrees that is defined in the discussion of EZW. Set Partitioning refers to the way these quadtrees divide up and partition the wavelet transform values at a given threshold. By a careful analysis of this partitioning of transform values, Said and Pearlman were able to develop the EZW algorithm, considerably increasing its compressive power. SPIHT algorithm produces an embedded bit stream from which the best reconstructed images with minimal mean square error can be extracted at various bit rates. Some of the best results like highest PSNR values for given compression ratios for a wide range of images, have been obtained with SPIHT algorithm.

The SPIHT method is not a simple extension of traditional methods for image compression, and represents an important advance in the field. The main features of SPIHT coding are:

• good quality reconstructed images, high PSNR, especially for colour images
• optimized for progressive image transmission
• produces a fully embedded coded file
• simple quantization algorithm
• fast coding/decoding (near symmetric)
• has wide applications, fully adaptive
• can be used for lossless compression
• can code to exact bit rate or distortion
• efficient combination with error protection

SPIHT coding yields all these qualities simultaneously. This makes it really outstanding. Table 2 shows the compression ratio and PSNR results for SPIHT algorithm.

Table 2. Results for Lena test image of size 256x256:
Compression Ratio & PSNR using SPIHT

| Level | Bit Planes Discarded | Compression Ratio | PSNR |
|---|---|---|---|
| 3 | 3 | 6.57 | 31.28 |
| 3 | 5 | 13.03 | 26.81 |
| 4 | 3 | 9.44 | 29.00 |
| 4 | 5 | 28.38 | 25.87 |
| 5 | 3 | 10.38 | 26.76 |
| 5 | 5 | 38.94 | 24.66 |

It can be seen that when the level of decomposition is increased, the compression ratio increases. This is because the coefficients with higher magnitudes concentrate mostly on the root levels as the decomposition level increases. Furthermore, most of the coefficients will have low magnitudes. These coefficients require only less number of bits for storage and transmission, thus increasing the compression ratio. But for higher decomposition levels, the resolution of the reconstructed image will reduce. The perceptual image quality, however, is not guaranteed to be optimal, as seen in Figure 2, since the coder is not designed to explicitly consider the human visual system (HVS) characteristics. Research has shown that there are three perceptually significant activity regions in an image: smooth, edge, and textured or detailed regions [31]. By incorporating the differing sensitivity of the human visual system to these regions in image compression schemes such as SPIHT, the perceptual quality of the images can be improved at all bit rates. Figure 2 shows the output of SPIHT coding for different decomposition levels.

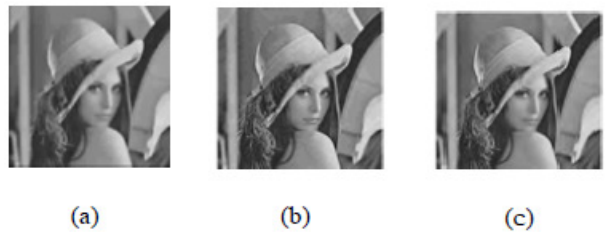

(a)  (b)  (c)

Figure 2. a) Original Image  b) decomposition level = 3, 0.7 bpp
c) decomposition level = 9, 0.1 bpp

In Figure 3, several SPIHT compressions of the Lena image are shown at different bit rates. The original Lena image is shown in Figure 3(i). Five SPIHT compressions are shown with compression ratios of 128:1, 64:1, 32:1, 16:1 and 8:1.

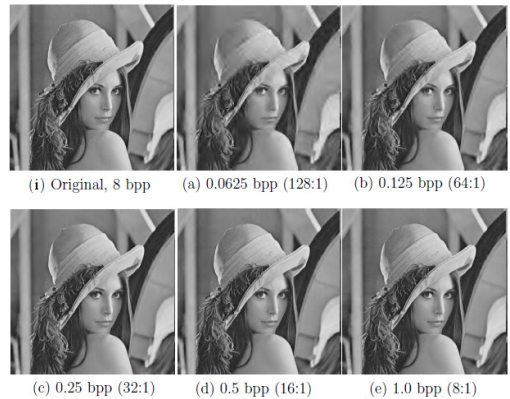

(i) Original, 8 bpp  (a) 0.0625 bpp (128:1)  (b) 0.125 bpp (64:1)

(c) 0.25 bpp (32:1)  (d) 0.5 bpp (16:1)  (e) 1.0 bpp (8:1)

Figure 3 : SPIHT compressions of Lena image. PSNR values: (a) 27.96 dB. (b) 30.85 dB. (c) 33.93 dB. (d) 37.09 dB. (e) 40.32 dB.

There are several things worth noting about these compressed images. First, they were all produced from one file, the file containing the 1 bpp compression of the Lena image. By specifying a bit budget, a certain bpp value up to 1, the SPIHT decompression will stop decoding the 1 bpp compressed file once the bit budget is exhausted. This illustrates the embedded nature of SPIHT. Second, the rapid convergence of the compressed images to the original is nothing short of astonishing. Even the 64: 1 compression in Figure 3(b) is almost indistinguishable from the original. A close examination of the two images is needed in order to see some differences, e.g., the blurring of details in the top of Lena's hat. Third, notice that the 1 bpp image has a 40:32 dB PSNR value and is virtually indistinguishable from the original even under very close examination. Effectiveness of the algorithm can be further enhanced by entropy coding its output, but at the cost of a larger encoding/decoding time. The need for reducing the number of bits used in this scheme led to the formation of the subsequent algorithm, called SPECK.

### 5.3. The SPECK Coder (Set Partitioned Embedded Block Coder)

The SPECK [12] algorithm can be said to belong to the class of scalar quantized significance testing schemes. The roots of this algorithm primarily lie in the ideas developed in the SPIHT, and a few block coding algorithms. The SPECK coder is different from the aforementioned schemes in that it does not use trees which span and also exploits the similarity across different sub bands. As an alternative, it makes use of sets in the form of blocks. The main idea is to exploit the clustering of energy in frequency and space in the hierarchical structures of wavelet transformed images.

The SPECK image coding scheme has all the properties that characterises the scalar quantized significance testing schemes. It shows the following properties in particular:

- Completely embedded: A certain coded bit stream can be used to decode the image at any rate less than or equal to the coded rate. It gives the finest reconstruction possible with the particular coding scheme.
- Employs progressive transmission: Source samples are encoded in decreasing order of their information content.
- Low computational complexity: The algorithm is very simple and does not require any complex computation.

- Fast encoding/decoding: due to the low computational complexity of the algorithm.
- Low dynamic memory requirements: During the coding process, at any given time, only one connected region, lying completely within a subband is processed. After processing this region, the next region is considered for processing.
- High efficiency: Its performance is comparable to the other low-complexity algorithms available today.

A brief discussion of the coding method of SPECK algorithm is given here. Consider an image F which has been adequately transformed using an appropriate subband transformation, for example, the most commonly used discrete wavelet transform. The transformed image exhibits a hierarchical pyramidal structure defined by the decomposition levels, with the topmost level being the root. The finest pixels lie at the bottom level of the pyramid while the coarsest pixels lie at the root level. The SPECK algorithm makes full use of rectangular regions of the image. These regions or sets are referred to as sets of type S and may be of varying dimensions. The dimension of a set S is influenced by the dimension of the original image and also depends on the subband level of the pyramidal structure at which the set lies. During the course of the algorithm, sets of various sizes will be formed, depending on the characteristics of pixels in the original set. A set of size 1 will consists of just one pixel. Sets of type I refer to the other type of sets used in the SPECK algorithm. These sets are obtained by chopping off a small square region from the top left portion of a larger region. Figure. 3 illustrates a typical set I.

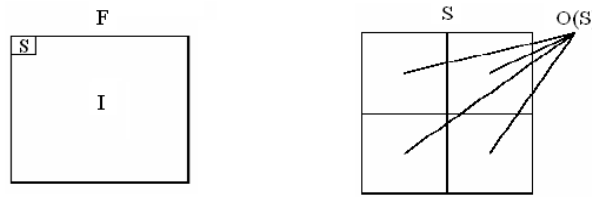

Figure 3. a) Partitioning of an image, b) Quadtree partitioning of set S
F into sets S & I,

Here two linked lists: LIS - List of Insignificant Sets, and LSP - List of Significant Pixels, are maintained. The LIS contains sets of type S of varying sizes which have not yet been found significant against a threshold n while LSP contains those pixels which have tested significant against n. Two types of set partitioning are used in SPECK: quad tree partitioning and octave band partitioning.

The idea of quad tree partitioning of sets, as shown in Figure 3, is to zoom in quickly to areas of high energy in the set S and code them first.

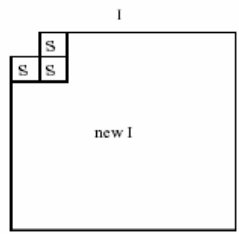

Figure. 4 Octave Partitioning

The idea behind octave band partitioning scheme shown in Figure 4, is to utilize the hierarchical pyramidal structure of the subband de-composition, where it is more likely that energy is concentrated at the top most levels of the pyramid. The energy content decreases gradually, as one goes down the pyramid. Partitioning a set S into four offsprings O(S) (i.e. forming sets S of a new reduced size) is the same as going down the pyramid one level at the corresponding finer resolution. Therefore, the size of a set S for a random image corresponds to a particular level of the pyramid.

The decoder uses the inverse mechanism of the encoder. During the execution of the algorithm, it takes significance test results from the coded bit stream and builds up the same list structure. Therefore, as the algorithm proceeds, it is able to follow the same execution paths for the significance tests of the different sets, and reconstructs the image progressively. It is observed that SPECK algorithm gives higher compression ratios. This has an advantage of processing sets in the form of blocks rather than in the form of spatial orientation trees. Table 3 gives the compression ratio and PSNR results for the same test image Lena using SPECK algorithm [12]. It is shown that images encoded with low decomposition levels are reconstructed with good resolution. For higher decomposition levels, the resolution is reduced marginally. Here most of the coefficients are of low magnitude. Hence, for a certain number of bit planes discarded, the percentage error for low magnitude is high. But for high magnitudes, it is negligible. So with smaller number of decomposition levels, the resolution of the reconstructed image can be made better than that in SPIHT as shown in Figure 6.

Table 3. Results for Lena test image of size 512x512:
Compression Ratio & PSNR using SPECK

| Level | Bitplanes Discarded | Compression Ratio | PSNR |
| --- | --- | --- | --- |
| 3 | 3 | 13.35 | 32.34 |
| 4 | 3 | 14.39 | 32.25 |
| 5 | 3 | 15.35 | 32.12 |
| 3 | 5 | 49.82 | 25.31 |
| 4 | 5 | 71.94 | 25.66 |
| 5 | 5 | 78.50 | 25.51 |

Figure. 5 shows reconstructed images of Lena using the SPECK algorithm for various values of decomposition levels and bitplanes discarded. The reconstructed images for the above compression ratios give appreciable resolution for the same decomposition levels and bitplanes discarded as compared with SPIHT, shown in Figure 6.

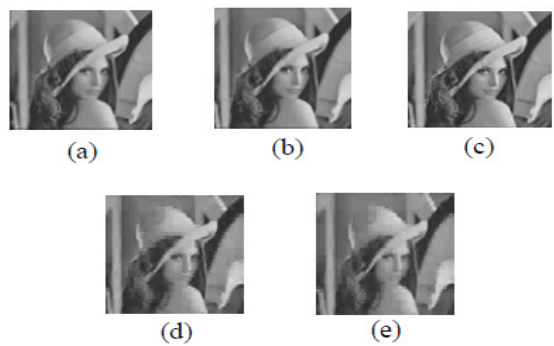

Figure 5. Reconstructed Images (a) Original Image (b) Bitplane discarded=3, decomposition level=3 (c) Bitplane discarded=3, decomposition level =5 (d) Bitplane discarded=5, decomposition level =4 (e) Bitplane discarded=5, decomposition level =5

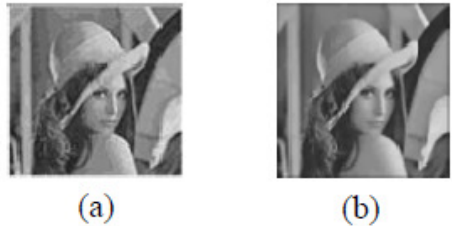

Figure 6. (a) SPIHT      (b) SPECK
Bitplane discarded = 3, decomposition level = 3

### 5.4. The EBCOT (Embedded Block Coding with Optimized Truncation) Algorithm

The EBCOT algorithm [13] shows advanced compression performance while producing a bit-stream with a rich set of features, like resolution and SNR scalability together with a "random access" property. All these features coexist within a single bit-stream without considerable reduction in compression efficiency. The EBCOT algorithm makes use of a wavelet transform to create the subband coefficients which are then quantized and coded. Although the usual dyadic wavelet decomposition is often used, other "packet" decompositions are also supported and occasionally preferred. The original image is characterized in terms of a collection of subbands, which may be organized into increasing resolution levels. The lowest resolution level comprises of the single LL subband. Each successive resolution level contains the additional subbands, that are required to reconstruct the image with twice the horizontal and vertical resolution.

The EBCOT algorithm is a scalable type image compression technique where the advantage is that the target bit-rate or reconstruction resolution need not be known at the time of compression. Another advantage of practical significance is that the image need not be compressed multiple times so as to achieve a target bit-rate, which is common in the existing JPEG compression technique. EBCOT algorithm divides each subband into comparatively small blocks of samples and creates a separate highly scalable bit-stream to represent each so called code-block.

The algorithm has modest complexity and is well suited to applications involving remote browsing of large compressed images. This algorithm uses code-blocks of size 64 x 64 with subblocks of size 16 x 16. The EBCOT bit-stream is composed of a collection of quality layers and SNR scalability is obtained by discarding unwanted layers.

The EBCOT images exhibit significantly less ringing around edges and superior interpretation of texture. Simulations have showed that some details preserved in the EBCOT images are totally lost by the SPIHT algorithm. However, the performance of EBCOT algorithm continues to be competitive with the state-of-the-art compression algorithms, considerably outdoing the SPIHT algorithm especially.

### 5.5. WDR (Wavelet Difference Reduction)

A major disadvantage of SPIHT algorithm is that it only implicitly locates the position of significant coefficients. Performing operations which depend on the location of significant transform values, such as region selection (a portion of a compressed image that requires increased resolution) on compressed data is a challenging task. For instance, this can occur with a portion of a low resolution medical image that has been sent at a low bit rate in order to arrive quickly. Such kind of compressed data operations are possible with the Wavelet Difference Reduction algorithm [14]. The term, difference reduction indicates the way in which WDR encodes the locations of the significant wavelet transform values. The main difference between WDR and bit-plane encoding is the significant pass. In Wavelet Difference Reduction, the output from the significance pass holds the signs of significant values along with sequences of bits which briefly describe the exact location of the significant values. Although WDR will not produce higher PSNR values than SPIHT method at low bit rates, as observed from Table 4, it can produce perceptually superior images, especially at high compression rates.

### 5.6. ASWDR (Adaptively Scanned Wavelet Difference Reduction)

An improvement in the WDR algorithm gave rise to the ASWDR algorithm [15]. This algorithm was introduced by Walker. The term adaptively scanned says that this algorithm modifies the scanning order used in WDR to achieve improved performance. ASWDR adapts the scanning order so as to forecast locations of new significant values. If a prediction is true, then the output specifying that location will just be the sign of the new significant value and the reduced binary expansion of the number of steps will be empty. So a good prediction scheme will considerably reduce the coding output of WDR. The forecast method used by ASWDR is as follows:

If w(m) is significant for threshold T, then the values of the children of m are expected to be significant for half threshold T/2. For several natural images, this prediction method is found rationally good. Table 4 shows the improved PSNR values for ASWDR compared to WDR.

Table 4. Performances (PSNR) of SPIHT, WDR & ASWDR on Lena

| Bit rate (bpp) | SPIHT | WDR | ASWDR |
| --- | --- | --- | --- |
| 0.5 bpp | 37.09 | 36.45 | 36.67 |
| 0.25 bpp | 33.85 | 33.39 | 33.64 |
| 0.125 bpp | 30.85 | 30.42 | 30.61 |

The WDR and ASWDR algorithms allow for ROI while SPIHT does not. In addition, multiresolution detection is facilitated by their superior performance in displaying edge details at low bit rates. Figure 7 shows magnifications of 128:1 and 64:1 compressions of the "Lena" image. The WDR compression does a better job in preserving the shape of Lena's nose and in retaining some of the striping in the band around her hat at 0.0625 bpp. Similar remarks apply to the 0.125 bpp compressions. However, the SPIHT algorithm preserves parts of Lena's eyes better. The ASWDR compressions yet better preserves the shape of Lena's nose and fine details of her hat, and show less distortion along the side of her left cheek and eyes (especially for the 0.125 bpp).

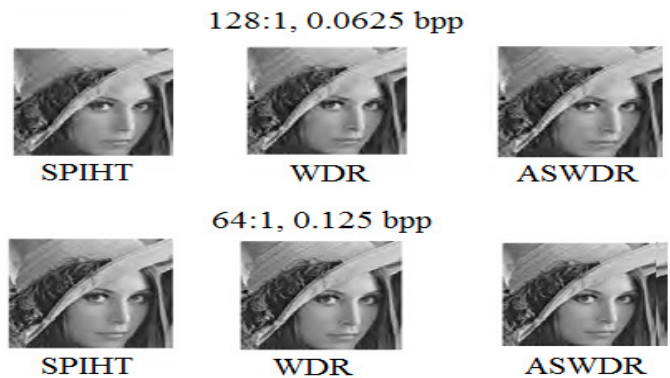

Figure 7. SPIHT, WDR, ASWDR Compressions

### 5.7. SFQ (Space - Frequency Quantization)

The standard separable 2-D wavelet transform has lately achieved great success in image processing because it provides a sparse representation of smooth images. However, it fails to efficiently capture 1-D discontinuities, like edges or contours. These features, being elongated and characterized by geometrical regularity along different directions, intersect and make many large magnitude wavelet coefficients. Since contours are very important elements in the visual perception of images, it is fundamental to preserve good reconstruction of these directional features to provide a good visual quality of compressed images. This is achieved by space-frequency quantization (SFQ) compression algorithm using directionlets [16].

SFQ [17] is a wavelet based image coding scheme that fits itself into a new class of image coding algorithms which combines standard scalar quantization of frequency coefficients with the tree structured quantization. The performance of the algorithm is very good and appears to confirm the promised efficiencies of hierarchical representation.

The Space - Frequency Quantization technique exploits both spatial and frequency compaction property of the wavelet transforms through the use of two simple quantization modes. It defines a symbol that indicates a spatial region of high frequency coefficients having zero value. This is to exploit the spatial compaction property. Application of this symbol is told as zero-tree quantization, because it involves setting to zero a tree-structured set of wavelet coefficients. This is done in the first phase which is called Tree Pruning Algorithm. In the next phase, called Predicting the Tree, the relation between a spatial region in image and the tree - structured set of coefficients is exploited. A mechanism for pointing to the location where high frequency coefficients are clustered is called Zero tree quantization. Thus, this quantization mode directly exploits the spatial clustering of high frequency coefficients predicted. For coefficients that are not set to zero by the zero tree quantization, a common uniform scalar quantization, independent of coefficient frequency band, is applied. Nearly optimal coding efficiency is provided by uniform scalar quantization followed by entropy coding.

### 5.8. The EPWIC (Embedded Predictive Wavelet Image Coder)

An embedded wavelet image coder based on statistical characterization of natural images in the wavelet transform domain is the EPWIC [18]. This technique defines the joint distribution between pairs of coefficients at adjacent spatial locations, orientations, and scales. Here, while the raw coefficients are nearly uncorrelated, their magnitudes are highly correlated. A reasonable description of the conditional probability densities is provided by a linear magnitude predictor together with both multiplicative and additive uncertainties. In EPWIC, using a non-adaptive arithmetic encoder, subband coefficients are encoded one bit-plane at a time. Bit-planes are ordered using an algorithm that considers the Mean Square Error reduction per encoded bit and the overall ordering is determined by the ratio of their encoded variance to compressed size. The predictive image coder proves useful in progressive transmission applications and is inherently embedded.

### 5.9. The CREW (Compression with Reversible Embedded Wavelet) Coding

A new form of wavelet transform based still image compression technique developed at the RICOH California Research Centre in Menlo Park, California, is the CREW [19]. It is both lossy and lossless compression scheme, bi-level and continuous-tone, progressive by resolution and pixel depth. The CREW can preserve the source image at encoder and quantize for the target device at decoder or transmission. It is hierarchical and progressive by nature.

Three new technologies combine to make CREW possible: the reversible wavelet transform (non-linear filters that have exact reconstruction implemented in minimal integer arithmetic), the embedded code stream (a method of implying quantization in the code stream) and a high-speed, high-compression binary entropy coder

The main features of CREW are:

- The same CREW code stream can be used for both lossless and lossy compression applications due to embedded quantization [26].
- The wavelet transform produces hierarchically ordered data and a natural means for interpolation.
- The bit-significance coding of CREW allows for bitplane progressive transmission.
- CREW compression is idempotent and CREW encoding and decoding can be efficiently implemented in a simple way either in hardware or software.

All these features combine to make a flexible "device-independent" compression system. CREW was the motivation for the new JPEG 2000 standard. The peculiar features make CREW [26] an ideal choice for applications that require high quality and flexibility for multiple input and output environments, such as, continuous-tone facsimile documents, medical imagery, fixed-rate and fixed-size applications (ATM), pre-press images, image archival, World Wide Web image or graphic type, satellite images etc. Not many of these applications have ever used compression schemes since the quality could not be assured, the compression rate was not high enough, or the data rate was not controllable.

### 5.10. The Stack- Run (SR)

SR [20] image coding algorithm works by the technique of raster scanning [24] within subbands. Therefore it involves much lower addressing complexity than other compression algorithms such as zero tree coding which requires creation and maintenance of lists of dependencies across different decomposition levels. The SR algorithm is a new approach in which a 4-ary arithmetic coder is used to represent significant coefficient values and the length of zero runs between coefficients. Despite its simplicity and uncomplicatedness and also the fact that these dependencies are not explicitly used, this algorithm is competitive with the finest enhancements of embedded zero tree wavelet coding.

### 5.11. The GW (Geometric Wavelet) Coding

Geometric wavelet is a recent advancement in the field of multivariate nonlinear piecewise polynomials approximation proposed by Deckel et.al [21]. It is a new and efficient method best suited for low bit-rate image coding. It combines the binary space partition tree approximation [23] with geometric wavelet (GW) coding so as to efficiently capture curve singularities and thus providing a sparse representation of the image. The GW method successfully competes with advanced wavelet methods such as the EZW, SPIHT, and EBCOT algorithms. A gain of about 0.4 dB over the SPIHT and EBCOT algorithms at the bit-rate 0.0625 bits-per-pixels (bpp) is achieved. The GW technique also outperforms other recent segmentation methods that are based on "sparse geometric representation". Table 5 shows the PSNR values for Lena test image using the GW approach.

Table 5.  PSNR values for Lena image of size 512x512 using GW, compared with other state-of-the-art methods

| Compression ratio | bit-rate (bpp) | GW | EZW | SPIHT | EBCOT |
|---|---|---|---|---|---|
| 1:64 | 0.125 | 30.74 | 30.23 | 31.10 | **31.22** |
| 1:128 | 0.0625 | **28.74** | 27.54 | 28.38 | 28.30 |
| 1:256 | 0.0315 | **26.64** | 25.38 | 26.1 | — |

An improvement over the geometric wavelet (GW) image coding method by using the slope intercept representation of the straight line in the binary space partition scheme is proposed by Chopra and Pal [22]. The performance of this algorithm is compared with other classical wavelet transform-based compression methods such as EZW, the SPIHT and the EBCOT algorithms. The proposed image compression algorithm outperforms the EZW and SPIHT algorithm, as shown in Table 6. The PSNR results for Lena image using the improved GW method for compression ratios of 64, 128 and 256 are given in Table 6. These results are also compared with other algorithms.

Table 6.  PSNR values for Lena image of size 512x512 using improved GW, compared with other state-of-the art-methods

| Method | Compression Ratio | | |
|---|---|---|---|
| | 256:1 | 128:1 | 64:1 |
| EZW | 25.38 | 27.54 | 30.23 |
| SPIHT | 26.10 | 28.38 | 31.10 |
| EBCOT | — | 28.30 | **31.22** |
| Bandelets | — | — | 30.63 |
| GW | 26.64 | 28.72 | 30.73 |
| Improved GW | **26.67** | **28.78** | 30.82 |

## 6. SUMMARY

Various features of the main coding schemes mentioned in the previous sections are briefed here. In Table 7, the features of various wavelet-based image coding techniques and their shortcomings are tabulated. It is seen that the latest techniques such as GW, EBCOT, and ASWDR are performing better than its predecessors such as EZW and WDR. Each technique has its own advantages and suits well with different images.

Table 7.  Various coding schemes: Performance and demerits

| TYPE | FEATURES | SHORTCOMINGS |
|---|---|---|
| *EZW* | • Employs progressive and embedded transmission<br>• Uses zerotree concept<br>• Uses predefined scanning order<br>• Good results without pre-stored tables, codebooks / training | • Transmission of coefficient position is missing<br>• No real compression<br>• Followed by arithmetic encoder |
| *SPIHT* | • Widely used<br>• high PSNR values for given CRs for variety of images<br>• Quad- tree or hierarchical trees are set | • Only implicitly locates position of significant coefficients<br>• More memory requirements<br>• Suits variety of natural images |
| | partitioned<br>• Employs progressive and embedded transmission<br>• Superior to JPEG in perceptual image quality and PSNR | • Perceptual quality not optimal |
| **SPECK** | • Does not use trees<br>• Uses rectangular block regions<br>• Employs progressive and embedded transmission<br>• Low computational complexity<br>• Low memory requirements<br>• Better PSNR than SPIHT | |
| **EBCOT** | • Supports packet decompositions also<br>• Block based scheme<br>• Modest complexity<br>• SNR scalability can be obtained<br>• Less ringing around edges<br>• Superior rendition of textures<br>• Preserves edges lost by SPIHT | • As layers increase, performance decreases<br>• Suits applications involving remote browsing of large compressed images |
| **WDR** | • Uses ROI concept<br>• Better perceptual image quality than SPIHT<br>• No searching through quad trees as in SPIHT<br>• Suits low resolution medical images at low bit rate<br>• Less complex than SPIHT<br>• Higher edge correlations than SPIHT<br>• Better preservation of edges than SPIHT | • PSNR not higher than SPIHT |
| **ASWDR** | • Modified scanning order compared to WDR<br>• Encodes more significant values than WDR<br>• PSNR better than SPIHT and WDR<br>• Perceptual image quality better than SPIHT and slightly better than WDR<br>• Slightly higher edge correlation values than WDR<br>• Preserves more of the fine details<br>• Suits high CR images like in reconnaissance and medical images | |
| **GW** | • Better PSNR compared to EZW and SPIHT | • High computational complexity<br>• High execution time |

Each of these schemes finds use in different applications owing to their unique characteristics. Though there a number of efficient coding schemes available, wide commercial usage and the need for improved performance demand the development of newer and better techniques.

# 7. CONCLUSION

The wavelet-based image compression has gained much popularity because of their overlapping nature that reduces the blocking artifacts. The multiresolution character of wavelet-based schemes which leads to superior energy compaction with high quality reconstructed images. A detailed survey on some of the popular wavelet coding techniques is summarized. Recommendations and discussions are presented for algorithm development and implementation.

## ACKNOWLEDGEMENT

The authors thank the editors and reviewers for their valuable comments and suggestions that helped us to make the paper in its present form.

## REFERENCES

[1] David Salomon, "Data Compression: The Complete Reference", Springer international Edition, 2005.
[2] Rafael C. Gonzalez and Richard E. Woods, "Digital Image Processing", Pearson Edition, 2005.
[3] G. K. Wallace, "The JPEG still-picture compression standard" Commun. ACM, vol. 34, pp. 30–44, Apr. 1991.
[4] MPEG-2video, ITU-T-Recommendation H.262-ISO/IEC 13818-2, Jan. 1995.
[5] K.P.Soman & K.I.Ramachandran "Insight into Wavelets from theory to practice", Prentice Hall India, New Delhi, 2002.
[6] M. Vetterli, and J. Kovacevic, "Wavelets and Subband Coding", in Englewood Cliffs, NJ, Prentice Hall, 1995, http://cm.bell-labs.com/ who/ jelena/Book/home.html
[7] Subhasis Saha, "Image compression - from DCT to Wavelets-AReview"http://www.acm.org/crossroads/xrds6-3/sahaimcoding.html
[8] Hiroshi Kondo and Yuriko Oishi, "Digital Image Compression using directional sub-block DCT".
[9] Jerome M. Shapiro, "Embedded Image coding using zerotrees of wavelets coefficients," IEEE Transactions on signal processing, vol. 41, no. 12, pp. 3445-3462, Dec 1993.
[10] Marc Antonini, Michel Barlaud, Pierre Mathieu & Ingrid Daubechies, "Image coding using wavelet transform", IEEE Transactions on image processing,Vol. 1 No. 2,April 1992.
[11] Asad Islam and W. Pearlman, "A. A New, Fast, and Efficient Image Codec Based on Set Partitioning in Hierarchical Trees", IEEE Trans. on Circuit & Systems for Video Technology, vol. 6, no. 3, pp. 243-250, June 1996.
[12] F. Khelifi, F. Kurugollu, A. Bouridane, "SPECK-Based Lossless Multispectral Image Coding ," IEEE Signal Processing Letters, vol. 15, pp. 69-72, Jan 2008.
[13] D. Taubman, "High Performance Scalable Image Compression with EBCOT," IEEE Trans. on. Image Processing, vol. 9, no.73, pp. 1158-1170, July 2000.
[14] J. S. Walker, T. Q. Nguyen, "Adaptive scanning methods for wavelet difference reduction in lossy image compression," in Proc. International Conf. Image Processing, 2000, pp. 182-185.
[15] Walker, J.S., "A Lossy Image Codec Based on Adaptively Scanned Wavelet Difference Reduction", Optical Engineering, in press.
[16] V Velisavljevic, B Beferull-Lozano, M Vetterli, "Space-Frequency Quantization for Image Compression With Directionlets ," IEEE Trans. Image Process., vol. 16, no. 7, pp. 1761-1773, July 2007.
[17] Xiong, Z., Ramachandran, K. and Orchard, M. T. Space-Frequency Quantization for Wavelet Image Coding, IEEE Trans. on Image processing, vol. 6, no.5, pp.677-693, May 1997.
[18] R. Buccigrossi, and E.P. Simoncelli, "EPWIC: Embedded Predictive Wavelet Image Coder," GRASP Laboratory, TR#414, http://www.cis.upenn.edu/~butch/EPWIC/index.html
[19] A. Zandi, J. D. Allen, E. L. Schwartz, M. Boliek, "CREW: Compression with Reversible Embedded Wavelets," in Proc. Data Compression Conf., 1995, pp. 212-221
[20] Min-Jen Tsai "Very low bit rate color image compression by using stack-run-end coding ," IEEE Trans. on Consumer Electronics, vol. 46, no. 2, pp. 368-374, May 2000.

[21] D. Alani, A. Averbuch, and S. Dekel, "Image coding with geometric wavelets," IEEE Trans. Image Process., vol. 16, no. 1, pp. 69–77, Jan. 2007.
[22] Chopra, G. Pal, A.K, "An Improved Image Compression Algorithm Using Binary Space Partition Scheme and Geometric Wavelets" IEEE Trans. Image Process., vol. 20, no. 1, pp. 270–275, Jan. 2011.
[23] H. Radha, M. Vetterli, and R. Leonardi, "Image compression using binary space partitioning trees," IEEE Trans. Image Process., vol. 5, no. 12, pp. 1610–1624, Dec. 1996.
[24] C. Pepin, P. Raffy, R. M. Gray, "Robust stack-run coding for low bit-rate image transmission over noisy channels ," IEEE Signal Processing Letters, vol. 9, no. 7, pp. 196-199, July 2002.
[25] Asad Islam and Pearlman, "An embedded and efficient low-complexity, hierarchical image coder", Visual Communication and Image processing '99 proceedings of SPIE., vol 3653, pp. 294-305, Jan 1999.
[26] Boliek, M., Gormish, M. J., Schwartz, E. L., and Keith, A. Next Generation Image Compression and Manipulation Using CREW, Proc. IEEE ICIP, 1997. http://www.crc.ricoh.com/CREW.
[27] James S Walker "Wavelet-Based Image Compression" The Transform and Data Compression Handbook Ed. K. R. Rao et al. Boca Raton, CRC Press LLC, 2001.
[28] Marc Antonini, Michel Barlaud, Pierre Mathieu & Ingrid Daubechies, "Image coding using wavelet transform", IEEE Transactions on image processing, Vol. 1 No. 2, April 1992.
[29] Rao, K. R.and Yip, P. Discrete Cosine Transforms -Algorithms, Advantages, Applications, Academic Press, 1990.
[30] K. Sayood, "Introduction to Data Compression", 2nd Ed., Academic Press, Morgan Kaufmann Publishers, 2000.
[31] A. Islam, W. A. Pearlman "",Proceeding of the SPIE VCIP '99, Vol 3653, 1999.

**Authors**

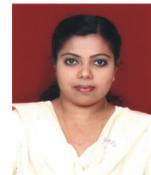

Rehna. V. J was born in Trivandrum, Kerala State, India on 1st Nonember 1980. She studied Electronics & Communication Engineering at the PET Engineering college, Vallioor, Tirunelveli District, Tamilnadu State, India fom 1999 to 2003. She received Bachelor's degree from Manonmanium Sundarnar University, Tirunelveli in 2003. She did post-graduation in Microwave and TV Engineering at the College of Engineering, Trivandrum and received the Master's degree from Kerala University, Kerala, India in 2005. Presently, she is a research scholar at the Department of Electronics and Communication Enginering, Noorul Islam Center for Higher Education, Noorul Islam University, Kumarakoil, Tamilnadu, India; working in the area of image processing under the supervision of Dr. M. K. Jeya Kumar. She is currently working as Assistant Professor at the Department of Telecommunication Engineering, Atria Institute of Technology, Bangalore, India. She has served as faculty in various reputed Engineering colleges in South India over the past seven years. She has presented and published a number of papers in national/international journals/conferences. She is a member of the International Association of Computer Science & Information Technology (IACSIT) since 2009. Her research interests include numerical computation, soft computing, image enhancement, coding and their applications in image processing.

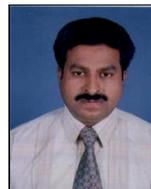

Dr. M. K. Jeya Kumar was born in Nagercoil, Tamilnadu, India on 18th September 1968. He received his Masters in Computer Applications degree from Bharathidasan University, Trichirappalli, Tamilnadu, India in 1993. He fetched his M.Tech degree in Computer Science and Engineering from Manonmaniam Sundarnar University, Tirunelveli, Tamilnadu, India in 2005. He completed his Ph.D degree in Computer Science and Engineering from Dr.M.G.R University, Chennai, Tamilnadu, India in 2010. He is working as a Professor in the Department of Computer Applications, Noorul Islam University, Kumaracoil, Tamilnadu, India since 1994. He has more than seventeen years of teaching experience in reputed Engineering colleges in India in the field of Computer Science and Applications. He has presented and published a number of papers in various national and international journals. His research interests include Mobile Ad Hoc Networks and network security, image processing and soft computing techniques.